\documentclass[journal]{IEEEtran}
\ifCLASSINFOpdf
\else
\fi

\usepackage{url}

\usepackage{amsmath}
\usepackage{graphicx}
\usepackage{multirow}
\usepackage{booktabs}

\usepackage{algorithmic}

\usepackage[linesnumbered,ruled,vlined]{algorithm2e}

\SetCommentSty{mycommfont}

\usepackage{xcolor}
\usepackage{tabularx}
\usepackage{multicol}
\usepackage{nicematrix}
\usepackage{array}
\usepackage{subcaption}
\usepackage{caption}
\usepackage{pifont}
\usepackage{soul}

\usepackage[most]{tcolorbox}

\begin{document}
\title{MC$^2$A: Enabling Algorithm-Hardware Co-Design for Efficient Markov Chain Monte Carlo Acceleration}

\author{Shirui~Zhao,~\IEEEmembership{Student Member,~IEEE,}
        Jun~Yin,~\IEEEmembership{Student Member,~IEEE,}
        Lingyun~Yao,~\IEEEmembership{Student Member,~IEEE,}
        Martin~Andraud,~\IEEEmembership{Member, IEEE}
        Wannes~Meert,~\IEEEmembership{Member,~IEEE},
        and~Marian~Verhelst,~\IEEEmembership{Fellow,~IEEE}

\thanks{Shirui Zhao, Jun Yin, and Marian Verhelst are with the Department of Electrical Engineering - MICAS, KU Leuven, Belgium (e-mail: firstname.lastname@kuleuven.be).}
\thanks{W. Meert is with the Department
  of Computer Science - DTAI, KU Leuven, Belgium (e-mail: firstname.lastname@kuleuven.be).}
\thanks{Lingyun Yao and Martin Andraud are with the Department of Electronics and Nanoengineering, Aalto University, 02150 Espoo, Finland (e-mail: firstname.lastname@aalto.fi).}
}

\maketitle
\vspace*{-8mm}
\begin{abstract}
An increasing number of applications are exploiting sampling-based algorithms for planning, optimization, and inference. The Markov Chain Monte Carlo (MCMC) algorithms form the computational backbone of this emerging
branch of machine learning. Unfortunately, the high computational cost limits their feasibility for large-scale problems and real-world applications. 
Several customized MCMC processors have been successful in accelerating such sampling-based computations. 
However, these solutions are either limited in hardware flexibility or fail to maintain efficiency at the system level across a variety of end-to-end applications. 

To overcome these hurdles, this paper introduces \textbf{MC$^2$A}, an algorithm-hardware co-design framework, enabling efficient and flexible optimization for MCMC acceleration. 
Firstly, \textbf{MC$^2$A}  analyzes the MCMC workload diversity through an extension of the processor performance roofline model with a 3rd dimension to derive the optimal balance between the compute, sampling and memory parameters within an MCMC processing hardware architecture.  
Secondly, \textbf{MC$^2$A} proposes a parametrized hardware accelerator architecture with flexible and efficient support of MCMC kernels with a pipeline of ISA-programmable tree-structured processing units, reconfigurable samplers and a crossbar interconnect to support irregular access.
This accelerator is programmable with a customized compiler, which maximizes parallelism, suppresses register/memory conflicts, and resolves pipeline hazards. 
Thirdly, the core of \textbf{MC$^2$A} is powered by a novel Gumbel sampler that eliminates exponential and normalization operations, increasing the throughput by $2\times$ without area overhead.

In the end-to-end case study, \textbf{MC$^2$A} achieves an overall {$307.6\times$, $1.4\times$, $2.0\times$, $84.2\times$} speedup compared to the CPU, GPU, TPU and state-of-the-art MCMC accelerator.
Evaluated on various representative MCMC workloads, this work demonstrates and exploits the feasibility of general hardware acceleration to popularize MCMC-based solutions in diverse application domains.
\end{abstract}


\begin{figure}[t]
    \centering
    \includegraphics[trim={0cm 0cm 0cm 0cm} , clip, width=0.8\columnwidth]{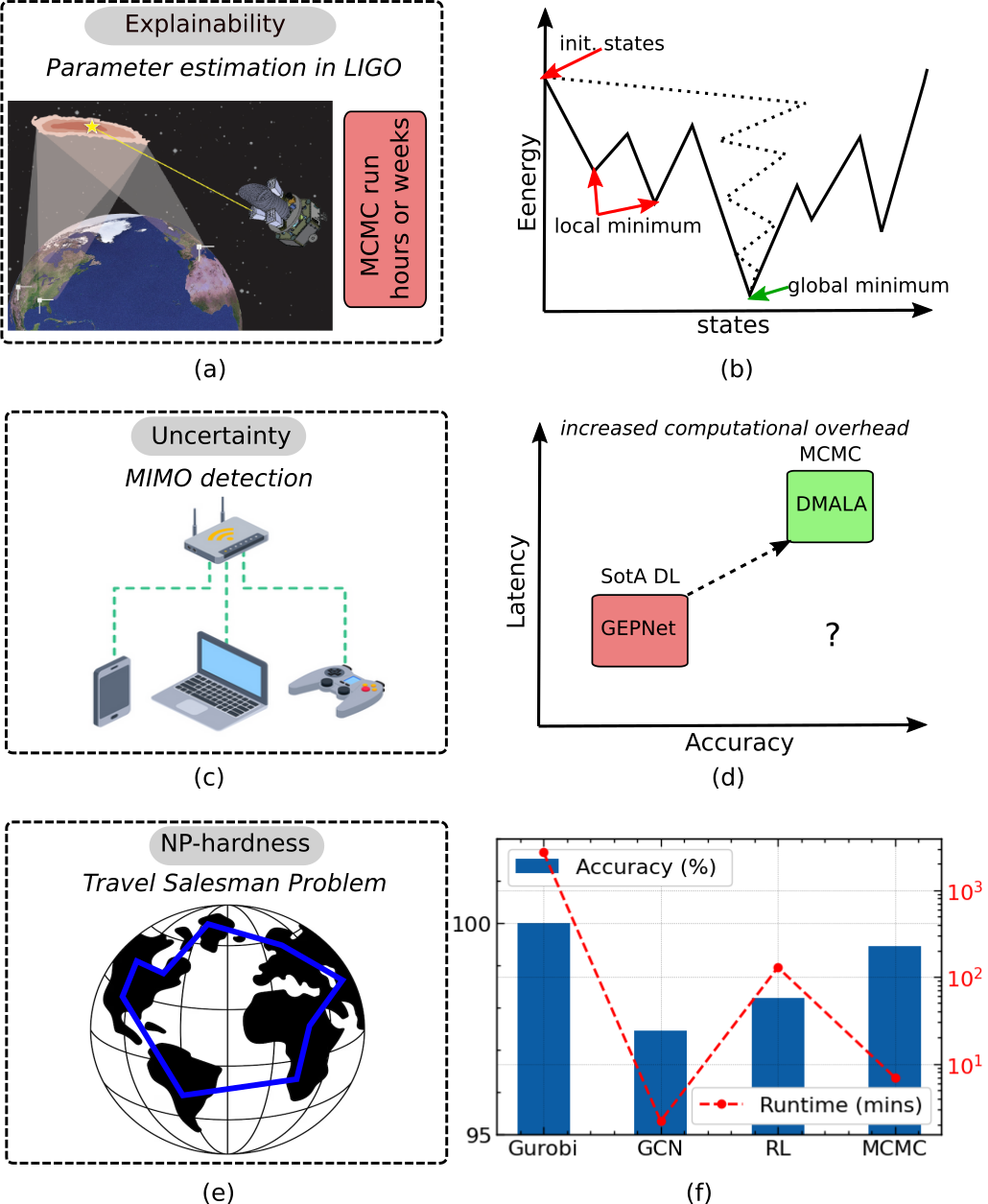}
    \caption{Markov Chain Monte Carlo (MCMC) methods are widely used to solve probabilistic problems across various domains. Real-world applications include: (a–b) parameter estimation in the LIGO system \cite{lalsuite}, which can require hours to weeks of computation; (c–d) MIMO detection \cite{zhou2024near}, where MCMC offers superior performance at the cost of increased computational complexity; and (e–f) COP such as the Traveling Salesman Problem \cite{sun23c}, where MCMC demonstrates promising results with higher accuracy and lower latency compared to other methods like Operations Research (e.g., Gurobi \cite{gurobi}) and Deep Neural Networks (DNNs) \cite{qiu2022dimes,li2018combinatorial,sun2023difusco}.
    } 
    \label{fig:mcmc_vs_nn}
\end{figure}

\section{Introduction}
Markov Chain Monte Carlo (MCMC) is recognized as one of the ten algorithms with the greatest influence on the development and practice of science and engineering in the 20th century \cite{top10}. 
MCMC methods are versatile and widely employed across various domains (Fig. \ref{fig:mcmc_vs_nn}), including Bayesian inference for complex models in probabilistic machine learning \cite{box2011bayesian, ghahramani2015probabilistic,andrieu2003introduction} (e.g., the Probabilistic Graphical Models (PGM) \cite{koller2009probabilistic}), intricate scientific computing \cite{liu2001monte} (e.g., Combinatorial Optimization Problems (COP) \cite{sun23c, katayoongoshvadiDISCSBenchmarkDiscrete2023}), and Energy-Based Models (EBM) \cite{lecun2006tutorial} (e.g., Restricted Boltzmann Machines (RBM)~\cite{bereux2024fast}). 
For example, MCMC algorithms are the pre-dominant techniques in the LIGO system to detect gravitational waves and estimate their physical parameters~\cite{christensen2022parameter}, which require several hours to weeks of computation depending on the implementation \cite{lalsuite} (Fig.~\ref{fig:mcmc_vs_nn}a). MCMC also show promising accuracy for the signal detection in multiple-input multiple-output (MIMO) systems with slightly high computational complexity~\cite{datta2013novel, bai2016large, zhou2024near}(Fig.~\ref{fig:mcmc_vs_nn}b-c). Many more COP-based applications (e.g., planning, logistics, manufacturing, etc.) can be solved by MCMC algorithms in an efficient way~\cite{sun23c}. 

In corresponding domains, MCMC-based methods can outperform alternative approaches such as Operations Research (OR) tools~\cite{gurobi} and Deep Neural Networks (DNNs)~\cite{sun23c,zhou2024near} (Fig. \ref{fig:mcmc_vs_nn}f). The advantage stems from MCMC's ability to efficiently sample from complex, multi-dimensional posterior distributions, enabling it to escape from local minimum optimazition results (Fig. \ref{fig:mcmc_vs_nn}b) and making it well-suited for probabilistic modeling and inference tasks that demand interpretability and reliability in noisy and uncertain environments~\cite{ghahramani2015probabilistic}. In contrast, DNNs excel in perceptual tasks $-$ such as image classification~\cite{krizhevsky2017imagenet} and natural language processing~\cite{chowdhary2020natural} $-$ where large datasets are available and posterior accuracy is less critical.

MCMC-based methods formalize an analytically intractable problem into a graph of random variables (RVs, i.e., collection of possible events) and efficiently solve it with a Markov chain that converges to a target distribution.
Though promising in the algorithmic performance, MCMC methods such as \textbf{Metropolis-Hastings (MH)} \cite{hastings1970monte} and \textbf{Gibbs sampling} \cite{gelfand1990illustration} suffer from their intrinsic iterative sampling procedure.
Over the past few decades, numerous MCMC variants have been developed to overcome this pitfall by exploiting parallelism potentials, especially in the handling of RVs.

\textbf{Block Gibbs sampling} \cite{jensen1995blocking} partitions the RVs into blocks based on the structure of graph, enabling simultaneous updates of RVs within the same block.
\textbf{Asynchronous Gibbs sampling} \cite{terenin2020asynchronous} allows RVs to be updated asynchronously, reducing the synchronization overhead and improving the sampling efficiency.
Advanced \textbf{gradient-based MCMC samplers} like the Path Auxiliary Sampler (PAS) \cite{pas_mcmc} and the Discrete Metropolis Adjusted Langevin Algorithm (DMALA) \cite{zhang2022langevin} have been proposed to leverage gradient information to compose new states, achieving better accuracy-latency tradeoffs than DNN methods (Fig. \ref{fig:mcmc_vs_nn}f).

Nevertheless, from the hardware perspective, these MCMC algorithms pose significant challenges 
beyond general matrix computation in their nature of probabilistic sampling, which incorporates randomness and stochastic decision-making.
To address such computational demands, domain-specific accelerators (DSA) have been proposed
to enable real-time processing, offering notable improvements in performance and energy efficiency, especially for edge-computing scenarios. Recent works include PGMA \cite{pgma}, CoopMC \cite{coopmc}, and PROCA \cite{proca}, which all leverage customized hardware to accelerate the MCMC sampling.

\begin{figure}[t]
    \centering
    \includegraphics[trim={0cm 0cm 0cm 0cm} , clip, width=\columnwidth]{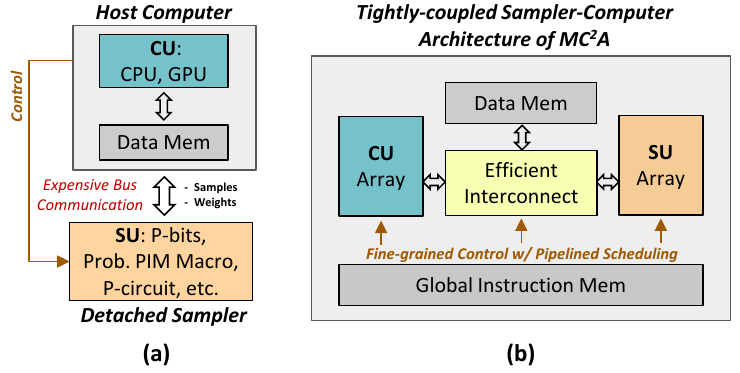}
    \caption{Comparison of MCMC accelerator structures: (a) The traditional system \cite{proca, li202512, chowdhury2023full} using decoupled Sample Unit (SU) and Compute Unit (CU) architecture with expensive inter-unit communication costs; (b) MC$^2$A's optimized architecture with a tightly-coupled sampler-computer collaboration and fine-grained pipeline control.
    }
    \label{fig:arch_idea}
\end{figure}

However, the existing accelerators come with several limitations. 
\textbf{\ding{172} Specialized Hardware-Algorithm Tuning:} Customized ASIC \cite{spu,pgma, aadit2022massively} or Ising machine \cite{10873406,mohseni2022ising} are implemented for specialized workloads, such as Markov Random Field with a structured graph and Ising model with simple flip randomness.  
\textbf{\ding{173} Partial Algorithm Consideration:} Other accelerators, such as the P-bits based design \cite{chowdhury2023full}, focus solely on the efficiency of the sampler. Then, the communication overhead between the hardware sampler and the classical computation core (Fig. \ref{fig:arch_idea}a) compromises the overall system performance.
\textbf{\ding{173} Rigid Sampler Implementation}: with the ever-growing MCMC algorithm variants, accelerators bound to a certain type of sampler structure show weaknesses in new applications. For instance, the sequential samplers in \cite{proca, acmc} are specially tweaked for Gibbs/MH and suffer from high-dimensional problems or correlated parameter spaces due to their component-wise update nature.
On top of these, to our knowledge, there \textbf{lacks a general hardware-algorithm co-design framework} in this field that can explore the best parameters for such sampler-computer heterogeneous architecture at design time, as well as support and optimizes for different MCMC-based algorithms at run time.

In this paper, we take the first steps to bridge these gaps by introducing \textbf{MC$^2$A}, a novel algorithm-hardware co-design framework, to efficiently design and optimize MCMC-related hardware acceleration. \textbf{MC$^2$A} can be integrated into other heterogeneous SoC designs as a loosely-coupled accelerator. The main contributions of this paper are:

\begin{itemize}
\item \textbf{End-to-end HW-SW Optimization Scheme} 
that exposes parallelism opportunities and key bottlenecks in the MCMC algorithm and the target accelerator with an extended 3D Roofline Model. 
Together with a customized hardware compiler, \textbf{MC$^2$A} enables a design space exploration to find optimal hardware configurations for various MCMC algorithms that orchestrate the sampling, computation, and memory activities.
\item \textbf{Flexible Hardware Architecture} 
that supports an end-to-end acceleration for MCMC-based computation with run-time reconfigurability. 
It incorporates a tightly-coupled sampling-computing topology with optimized data path to efficiently handle the irregular dataflow patterns across different algorithms (Fig. \ref{fig:arch_idea}b).

\item \textbf{Novel Gumbel Sampler}
core that avoids expensive exponential and normalization operations to sample from unnormalized energy distributions. 
It also implements a reconfigurable parallelization scheme for different workloads.
\end{itemize}

The rest of the paper is organized as follows: 
Section \ref{sec:background} provides background knowledge of this work. 
Section \ref{sec:challenges_profiling} categorizes the MCMC accelerator design challenges that motivate our work.
Section \ref{sec:roofline} presents our 3D MCMC Roofline model, a powerful tool for rapid SW-HW profiling and design iteration.
Section \ref{sec:architecture} details the flexible and programmable \textbf{MC$^2$A} accelerator architecture with novel reconfigurable Gumbel sampler. 
Section \ref{sec:result} evaluates the performance of the MC$^2$A framework and conducts SoTA comparisons.
Finally, Section \ref{sec:conclusion} concludes the paper and outlines future research directions.

\section{Background and Foundational Algorithms} \label{sec:background}

In this section, we first introduce key variants of the MCMC algorithm (Section \ref{sec:background_baseline_mcmc}) and several representative applications (Section \ref{sec:background_mcmc_applications}) that require sampling-based approximate algorithms.
Then, we give a summary of required operators in MCMC processing from the hardware perspective (Section \ref{sec:background_required_op}).
Finally, we discuss the algorithmic potentials of MCMC for parallel processing (Section \ref{sec:background_parallelism_opportunities}).
These backgrounds lay the foundation of our proposed methodologies and benchmark evaluations.

Throughout this paper, we use capitalized $\textbf{X}$ to represent a set of random variables (RVs) and lowercase $\textbf{x}$ to represent the set of samples on the corresponding RVs. 
To simplify, we denote the probability distribution as $\textbf{P(x)}$, i.e. $\textbf{P(x)} = \textbf{P}(X_0 = x_0, X_1 = x_1, ...)$. 
Therefore, the energy function (the negative logarithm of the probability distribution function) is $\textbf{E(x)} = -\log \textbf{P(x)}$.


\subsection{Baseline MCMC Algorithms} \label{sec:background_baseline_mcmc}
\begin{algorithm}[!t]
\DontPrintSemicolon
\SetAlgoLined
\SetNoFillComment
\LinesNotNumbered
\caption{General MCMC Algorithm}
\label{alg:metropolis-hastings}
\tcc{\textbf{Chain level parallelism}}
\For{$i \gets 1, \dots, C$}{
    \tcc{\textbf{\emph{No parallelism, step $t$ depends on previous steps}}}
    \For{$t \gets 1, \dots, T$}{
        \tcc{\textbf{RV and computational level parallelsim}}
        Sampling \(x \sim Q(x|x_{t-1}) \) 
                
        \tcc{Accept or reject based on the probability}
        \(
        \alpha = \min \left( 1, \frac{P(x)}{P(x_{t-1})}\frac{Q(x_{t-1}|x)}{Q(x|x_{t-1})} \right)
        \)\;        
        \( x_t \gets \left\{
        \begin{array}{ll}
        x & \text{with probability } \alpha \\
        x_{t-1} & \text{with probability } 1-\alpha
        \end{array}
        \right. \)\;
    }
}
\end{algorithm}

The sampling in discrete space can be formulated as the following problem: find a possible set $\textbf{x}$ in a finite discrete space $\boldsymbol{\chi}$ to minimize the energy function. The target distribution:
\begin{equation}
    P(x) = \frac{exp(-\beta E(x))}{Z}, x\in \boldsymbol{\chi}
\end{equation}

with the inverse temperature $\beta$ as simulated annealing factor\cite{van1987simulated}, $E(x)$ as the energy function, and $Z$ as the normalization factor ($Z = \sum exp(-\beta E(z)), z \in \chi$). Here, the parameter $\beta$ serves as an alternative way to express the temperature $T$, where $\beta = \frac{1}{T}$.
However, the direct computing in high-dimensional spaces or upon the intractable normalized $Z$ is challenging \cite{ghahramani2015probabilistic}. To overcome such difficulty,
the following Markov Chain Monte Carlo (MCMC) methods are proposed.

\begin{figure*}[!t]
    \centering
    \includegraphics[width=1.0\textwidth]{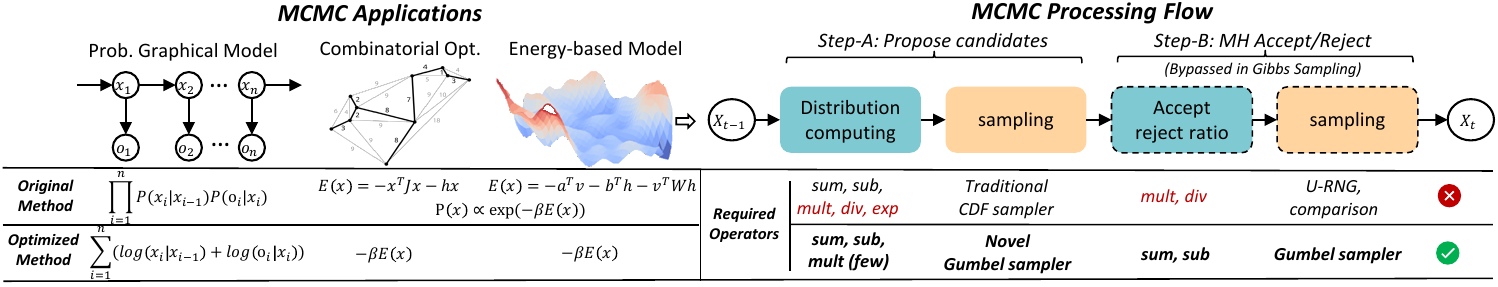}
    \caption{
    The overview of the MCMC processing flow, in view of (a) the key energy functions in different MCMC applications and (b) the required operators in the MCMC hardware. Hereby, we cross-compare the original approach and our optimized approach (Gumbel sampler, detailed in Section \ref{sec:architecture_se}). 
    }
    \label{fig:algorithm_flow}
\end{figure*}
\textbf{Sequential Metropolis-Hastings (MH) and Gibbs sampling.} 
The MH algorithm \cite{hastings1970monte} is a widely used MCMC method that generates a sequence of samples $x$ from a probability distribution. It works by proposing a new set of samples based on the previous steps and accepting or rejecting the new samples based on an acceptance probability that ensures a detailed balance (shown in Alg.~\ref{alg:metropolis-hastings}). The samples construct a Markov chain that converges to a target distribution $P(x)$. The MH algorithm is a general-purpose MCMC algorithm that can be applied to any distribution, but it may suffer from slow convergence \cite{katayoongoshvadiDISCSBenchmarkDiscrete2023}.
\textbf{Gibbs sampling} is a special case of the MH algorithm where the accept/reject ratio $\alpha$ always equals one, which is simpler and will not reject samples\cite{ghahramani2015probabilistic}. These two algorithms will only change the state of one RV per iteration.

\begin{figure}[!t]
    \centering
    \includegraphics[width=1.0\columnwidth]{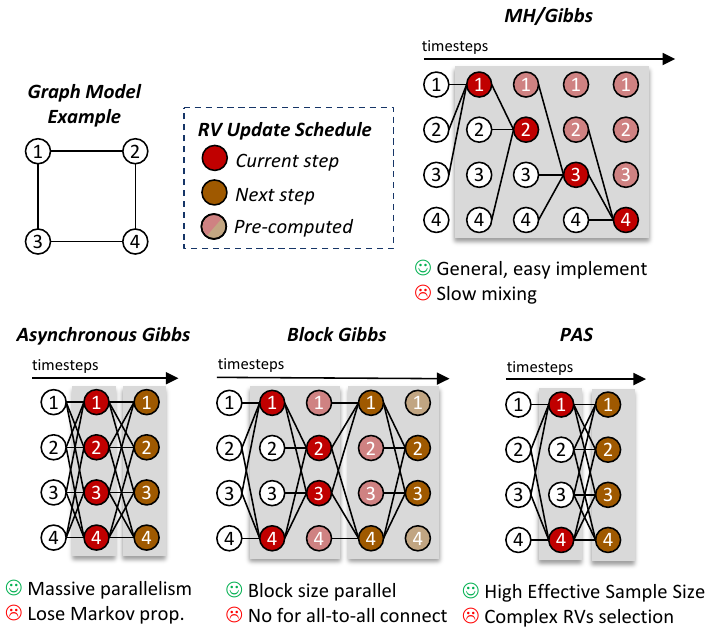}
    \caption{Illustration of the Random Variable (RV) level parallelism in different MCMC algorithms.}
    \label{fig:parallelism}
\end{figure}

\textbf{Block Gibbs (BG) sampling.}
Block Gibbs sampling is a parallel version of the Gibbs sampling algorithm that simultaneously updates multiple random variables (RVs) \cite{jensen1995blocking}.
It partitions the RVs into blocks based on the structure of the graphical model, allowing for simultaneous updates of RVs within the same block. Block Gibbs uses the principle of the \texttt{Markov Blanket} to detect the blocks. The \texttt{Markov Blanket} is a set of nodes that are conditionally independent of the rest of the graph given the other nodes. For example, in the 4-node undirected graph shown in Fig.~\ref{fig:parallelism}, the nodes in the Markov Blanket of node $1$ are $2$ and $3$, which means nodes $1$ and $4$ are conditionally independent of each other given $2$ and $3$. Thus, Block Gibbs can update the RVs of $1$ and $4$ simultaneously.

\textbf{Asynchronous Gibbs (AG) sampling.}
Asynchronous Gibbs sampling \cite{patelPASSAsynchronousProbabilistic2024,terenin2020asynchronous} allows RVs to be updated asynchronously at the same time step, improving the sampling efficiency. However, the asynchronous nature of this algorithm may lead to non-Markovian behavior, which can affect the convergence properties of the Markov chain.

\textbf{Gradient-based discrete sampling.} 
The most recent Path Auxiliary Sampler (PAS) \cite{sun23c} is an advanced gradient-based discrete sampling algorithm that outperforms other MCMC algorithms \cite{katayoongoshvadiDISCSBenchmarkDiscrete2023}. It introduces auxiliary path variables to facilitate parallel sampling for the most "dynamic" variables, which have the largest influence on the energy function. Each iteration of the PAS algorithm updates with an MH step as follows:
\begin{enumerate}
    \item \textbf{Find $L$ most "dynamic" variables:} Sample the indices of those random variables $\mathcal{J}=\{j_1,\dots,j_L\}$ from Categorical distribution ($\Delta E$) with size of $N$, where
    \begin{equation}
        \Delta E_i = \sum_{s \neq x_i} {E(x_{\backslash i}, X_i = s) - E(x)}, {i \in [0,N)}
    \end{equation}
    \item For $j \in \mathcal{J}$, sample $x_j \sim q_x^j(s)$, where:
    \begin{equation}
        q_x^j(s) = \frac{\exp(-\beta E(x_{\backslash j}, X_j = s))}{Z}
    \end{equation} 
    \item Accept $x$ with probability (same MH step shown in Alg.~\ref{alg:metropolis-hastings}).
\end{enumerate}
Where $L$ is a predefined hyperparameter indicating how many RVs can be updated in the current iteration step, $N$ is the total RV size, and $s$ represents the possible values for RVs $X$.

A key advantage of these gradient-based samplers is their simplicity and efficiency \cite{sun2023discrete}. They require no training and maintain a lighter computational footprint compared to machine learning alternatives. Compared to Gibbs sampling, gradient-based methods utilize full gradient descent, which could be faster but require more computational resources for gradient calculations.

\textbf{No Free Lunch theorem.}
According to the \textit{No Free Lunch} theorem \cite{wolpert1997no}, there is {\em no one-size-fits-all} algorithm that works best for all problems. This is especially true in the MCMC field \cite{katayoongoshvadiDISCSBenchmarkDiscrete2023}. 
The choice of MCMC algorithm depends on the problem structure, the computational resources available, and the desired trade-offs between speed and quality. For example, MH sampling is more general and easier to implement, but it may suffer from slow convergence speed in high-dimensional or correlated parameter spaces \cite{sun2023discrete}. On the contrary, gradient-based samplers, such as PAS \cite{sun23c}, can achieve faster convergence, but they require more computational resources for gradient calculations (Section \ref{sec:challenges_profiling}). 

It will in practice be important to support multiple algorithms, as it depends on the application which one is preferred.

\subsection{Representative MCMC Applications} \label{sec:background_mcmc_applications}
MCMC algorithms are widely used in machine learning and scientific computing. Here, we introduce several representative applications on how the MCMC methods in Section \ref{sec:background_baseline_mcmc} are leveraged.

\textbf{Probabilistic Graphical Models (PGM).}
PGM are a powerful tool in machine learning for representing complex relationships between variables in a probabilistic model. They are widely used in causal learning \cite{bitadarvishrouhaniCausaLearnAutomatedFramework2018}, computer vision \cite{yufeiniPMBAParallelMCMC2021}, and natural language processing \cite{pgma}.
\textbf{Bayesian Networks (Bayes Nets)} 
are PGM represented by directed acyclic graphs, 
a special case of Hidden Markov Models (HMM) shown in Fig.~\ref{fig:algorithm_flow}. 
Following the Bayesian rule, the joint probability of the whole graph is the product of all RVs' conditional probabilities.
The logarithm probability function is commonly used to reduce the computational cost, which can convert multiplication and division to addition and subtraction.

\textbf{Markov Random Fields (MRF) or Potts Model}, \textbf{Combinatorial Optimization Problems (COP)} and \textbf{Energy-based Model} are leveraging MCMC algorithms to find the optimal configuration $x$ that minimizes the total energy $E(x)$. 
For example, the image segmentation application can be represented by a 2D-grid MRF or Ising model (when label size is 2). Each segmentation label RV could have five neighbors (its associated image pixel and four neighbor RVs) with a possible label value of $l \in [0, L)$. Its energy function can be computed 
by assuming $x_i$ is equal to $l$, which will form a Categorial distribution with size of $L$. 
The corresponding energy computations are listed in Fig.~\ref{fig:algorithm_flow}, which shows the requirement of matrix-vector computation and exponential operation to convert energy to probabilistic distribution with the inverse temperature factor $\beta$.

\subsection{Required Operators in MCMC Processing} \label{sec:background_required_op}
For the previously mentioned applications, various kinds of MCMC algorithms (discussed in Sec.\ref{sec:background_baseline_mcmc}) can be used. We discussed the required operations based on the general form of MCMC (Alg.~\ref{alg:metropolis-hastings}). Three major steps are involved:
\begin{enumerate}
    \item {\bf Distribution computing:} compute the energy distribution $E(x)$ and the associated probability distribution $P(x)$ according to the equation shown in Fig.~\ref{fig:algorithm_flow}, energy function instead of probability function can be used for (1) avoiding computation underflow and overflow, (2) replacing costly multiplication and division to addition and substation, however, it requires exponential function to convert the energy to probability and normalize the distribution before sampling.
    \item  {\bf Distribution sampling:} get the samples by sampling from the distribution $q(x|x_{t-1})$ with pseudo-random number sampling algorithms like inverse transform sampling.
    \item {\bf (Optional) MH step:} accept or reject the samplings with probability of $\alpha$.
\end{enumerate}
{\bf MC$^2$A} optimized the computation flow by leveraging the \textit{Gumbel} sampling algorithm, which can sample directly from the un-normalized energy function to simplify the {\bf Distribution computing} without the \textit{Exponential} and \textit{normalization} operations. More details will be discussed in Sec.~\ref{sec:architecture_sampler}.

\subsection{Parallelism in MCMC}\label{sec:background_parallelism_opportunities}
Although the iterative sampling procedure of basic MCMC algorithms undermines the capability of parallel processing, there still exists several opportunities from the perspective of the entire workflow. As shown in Algorithm \ref{alg:metropolis-hastings}, two major chances are the chain-level parallelism and the RV/computational-level parallelism.

\textbf{Chain-level parallelism.}
Different chains (first loop in Alg.~\ref{alg:metropolis-hastings}) are independent and can be run on parallel hardware architectures, as found in modern parallel CPU/GPU hardware platforms \cite{sountsov2024running, tran2018simple,bingham2019pyro}. However, increasing the number of chains is not always desirable, as it does not guarantee faster convergence for all workloads~\cite{sun23c}.
Therefore, this paper focuses on single-chain acceleration, while all presented techniques of MC$^2$A can still be easily scaled to support multiple chains by adding SIMD or instantiating multiple parallel MC$^2$A cores.

\textbf{RV-level parallelism.}
In each MCMC chain, all random variables (RVs) must be updated iteratively. Across different iterations (the second loop in Alg.~\ref{alg:metropolis-hastings}), there is no opportunity for parallelism, since the RVs at the current time step depend on the values from the previous step. However, within a single iteration, parallelism can still be exploited depending on the nature of the MCMC algorithm, as shown in Fig.~\ref{fig:parallelism} and explained below:
\begin{itemize}
    \item \textit{MH and Gibbs Sampling} are inherently sequential. Each RV is updated one at a time in a fixed order. For example, RVs 0–4 are updated sequentially within one iteration, limiting opportunities for parallel execution.
    \item \textit{Block Gibbs Sampling} exploits conditional independence by grouping RVs into blocks. RVs within the same block can be updated simultaneously. In the given example, the RVs are partitioned into two blocks—(RV1, RV4) and (RV2, RV3)—each of which can be updated in parallel.
    \item \textit{Asynchronous Gibbs Sampling} updates all RVs concurrently without waiting for updates from other RVs. While this allows for full parallelism (e.g., RVs 0–4 are updated simultaneously), it breaks the strict Markov dependency structure, which may affect theoretical convergence guarantees.
    \item \textit{Gradient-based Samplers (e.g., PAS)} rank RVs based on their "dynamism" (i.e., how much they change) and update the most dynamic ones in parallel. The set of RVs selected for update can vary at each step. In the example, RV1 and RV4 are updated in the current iteration, while all RVs may be updated in the next.
\end{itemize}
\textbf{Computational-level parallelism.}
To update each RV, it requires both {\em distribution computing} and {\em distribution sampling} (shown in Sec.~\ref{sec:background_required_op}). The {\em distribution computing} involves multiple rounds of computation to form the distribution.
Computational-level parallelism can be applied to accelerate the processing. For RV $X_i$, $O(\mathcal{N})$ complexity is required to compute the energy function $E(x_i|x)$, where $\mathcal{N}$ is the number of possible values of $x_i$. To fully utilize this parallelism, the neighboring nodes for a given node shall be able to access simultaneously.

\begin{figure}[!t]
    \centering
    \includegraphics[width=\columnwidth]{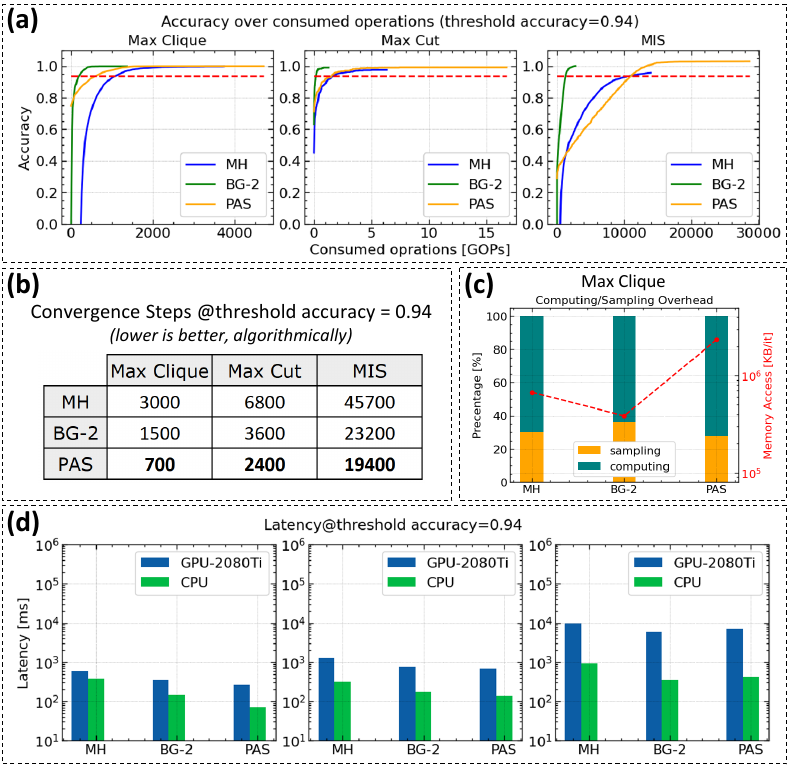}
    \caption{
    The demonstration of MCMC hardware challenges over different problems (Max Clique, Max Cut and MIS) and algorithms (MH, BG-2, PAS). With a target accuracy of 0.94 as threshold, we profile:  the different overheads measured by consumed operations (a) and algorithmic steps (b); (c) the Max Clique hardware overhead in terms of the computing/sampling ratio and memory access; (d) the different latencies on CPU and Nvidia GPU. 
    }
    \label{fig:latency_profile}
\end{figure}

\section{Observed Hardware Challenges of MCMC}\label{sec:challenges_profiling}

From the \textit{No-Free-Lunch} discussion in Section \ref{sec:background_baseline_mcmc}, MCMC hardware acceleration should be flexible enough to support different MCMC algorithms. However, with the upsurge of these MCMC applications, the challenges of efficient MCMC processing on hardware become critical. 
In this section, we break down the current limitations in this field with hardware profiling results and the nature of MCMC algorithms. These findings motivate our research on \textbf{MC$^2$A}.

Despite the parallel optimizations in the algorithmic level (Section \ref{sec:background_parallelism_opportunities}), the MCMC computation flow is still challenging for modern hardware processors in the following aspects:

\textbf{Computational Complexity.}
MCMC algorithms are computationally intensive, requiring numerous iterations to converge to the target distribution. Each iteration involves generating distributions, sampling from a proposal distribution, computing the acceptance probability, and updating the state. These operations include reduction sums, multiplications, Random Number Generation (RNG), and sampling algorithms. Single-precision computation is necessary to support general applications. As discussed in \cite{sountsov2024running}, underflow and rounding errors can significantly impact the performance of certain applications. {\bf Log-domain single-precision floating-point} computation is needed to avoid these issues.

\begin{figure*}[t]
    \centering
    \includegraphics[width=\textwidth]{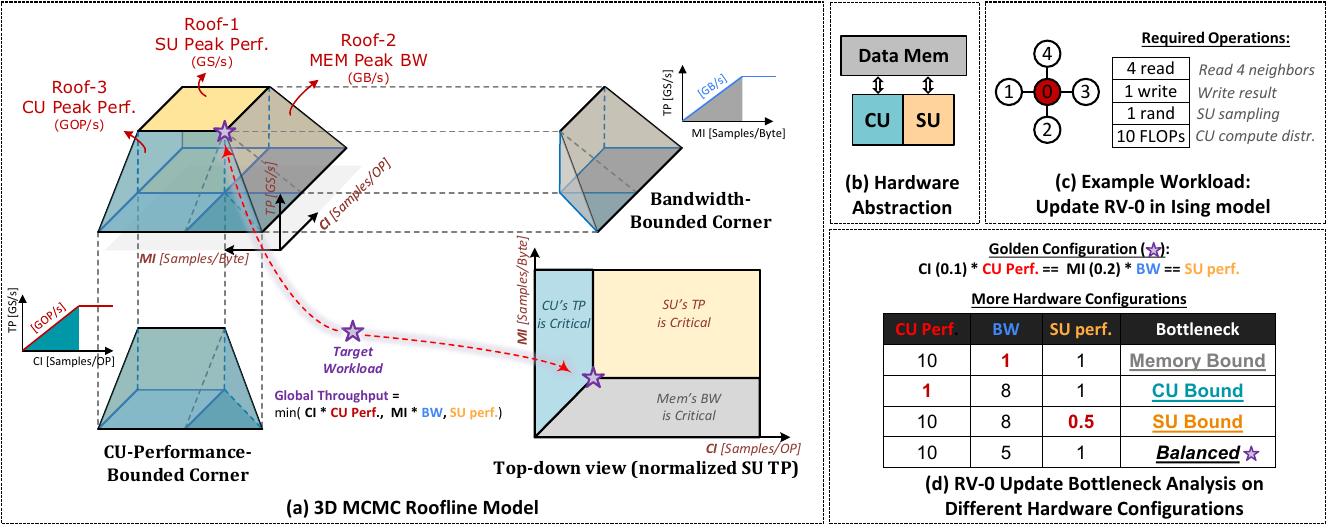}
    \caption{
    MC$^2$A 3D Roofline Model: (a) The 3D rectangular frustum that models the hybrid MCMC CU-SU hardware system. The peak hardware performances and their mutual data dependencies form three different performance boundary roofs and bottleneck corners. The purple star highlights the golden configuration where the algorithm and hardware reach a perfect match. (b) The CU-SU hardware abstraction we use. (c) The example workload showcases the usage of this 3D roofline. (d) The Bottlenck analysis of this workload on different hardware configurations. Abbreviations: CU = Compute Unit, SU = Sample Unit, CI = Computation Intensity, MI = Memory Intensity, TP = Throughput Performance, BW = Bandwidth, GS = GigaSamples.
    }
    \label{fig:3d_roofline}
\end{figure*}
\textbf{Hardware Inefficiency.}
Fig.~\ref{fig:latency_profile} presents the end-to-end accuracy across different time steps for three COP workloads \cite{katayoongoshvadiDISCSBenchmarkDiscrete2023} executed using \textbf{MG}, \textbf{BG-2} (Block Gibbs with block size of 2) and \textbf{PAS} sampling, respectively. As not all algorithms require the same amount of work, it is also interesting to look at the \textbf{\textit{consumed operations}} to reach a target accuracy, as shown in Fig.~\ref{fig:latency_profile}(a). 
The algorithms' JAX implementation allows to easily map the code on both CPU and GPU compute platforms. So we also measure the corresponding \textbf{\textit{latency performance}} on an Intel Xeon CPU and an Nvidia RTX GPU, which is shown in Fig.~\ref{fig:latency_profile}(d). Key observations include:
\begin{enumerate}
    \item From the comparison between Fig.\ref{fig:latency_profile}(a) and (b), advanced gradient-based algorithms (e.g., PAS sampler) reduce the number of iteration steps but might consume more operations in total.
    \item GPU latency performance is worse than the CPU, despite having more parallel hardware (Fig.~\ref{fig:latency_profile}(d)). 
    The reason are: i) low hardware utilization during reduce-sum operations for energy computation; ii) the bottleneck of sequential sampling operations ($\sim30\%$ of the total overhead, Fig.~\ref{fig:latency_profile}(c)). 
    \item We also notice that the BG-2 sampler suffers from the scaling up of problem dimensions when solving the Potts Model, which indicates an insufficient parallelization in its hardware implementation.
\end{enumerate}

\textbf{Memory Overhead.}
Figure \ref{fig:latency_profile}(c) also illustrates the memory access of the MaxClique workload measured by JAX profiling tools. We observe that MCMC algorithms consume GB-level memory in each iteration, primarily due to the large number of operations that rely on intermediate results and sampling processes to explore distribution items. 
Especially, the gradient-based samplers, such as the PAS, incur a higher amount of memory access to compute gradient information.
Such high memory requirements impact the efficiency of traditional hardware architectures, leading to memory-bounded bottlenecks.

\section{3D MCMC Roofline Model} \label{sec:roofline}
To design an optimal accelerator for MCMC, profiling tools are essential to pinpoint the bottlenecks in the system. 
From Fig. \ref{fig:arch_idea}, a typical MCMC processor should include a Sample Unit (\textbf{SU}) for sampling and a Compute Unit (\textbf{CU}) for energy computation.
Yet, the traditional processor performance roofline model is not sufficient to analyze this sampler-computer hybrid system, leading to Fig. \ref{fig:latency_profile}.
Therefore, \textbf{MC$^2$A} adapts a 3D roofline extension to reveal potential challenges in Section \ref{sec:challenges_profiling} and derive the optimal balance between the compute, sampling and memory parameters. 

Fig.~\ref{fig:3d_roofline} explains the details of our \textbf{MC$^2$A} roofline model. We take a hardware abstraction as Fig. \ref{fig:3d_roofline}(b) for the top-level modeling of our target hardware in Fig. \ref{fig:arch_idea}(b). Because the rest of the hardware constraints, such as the CU-SU internal communication, can be addressed during the tightly-coupled hardware design (Section \ref{sec:architecture}).

Our 3D roofline model is modeled from the perspective of SU performance. There are three key dimensions: Compute Intensity (\textbf{CI}), Memory Intensity (\textbf{MI}), and Throughput Performance (\textbf{TP}): 
\begin{itemize}
    \item \textbf{Computation Intensity (CI):} the ratio of SU sampling to the CU's \textit{Distribution computing} operations, in Samples/OP.  
    \item \textbf{Memory Intensity (MI):} the ratio of SU sampling to the memory's total data movement, in Samples/Byte. This includes all the memory fetches by the CU and SU to complete one sampling step.
    \item \textbf{Throughput Performance (TP):} the overall performance of the MCMC system, in Giga-samples per second ([GS/s]).
\end{itemize}

\begin{figure*}[!t]
    \centering
    \includegraphics[width=1.0\textwidth]{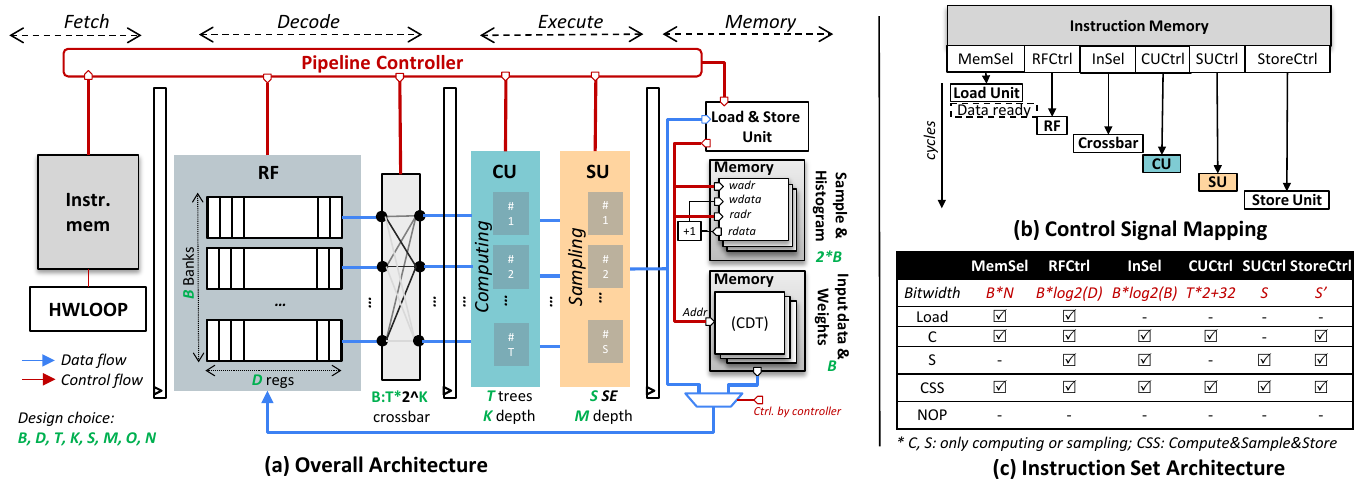}
    \caption{The overview of \textbf{MC$^2$A} top-level architecture: (a) The 4-stage pipelined and parameterizable architecture for tightly-coupled computing-sampling MCMC acceleration. (b) The pipelined mapping of control signals to each hardware modules. (c) The definition of the \textbf{MC$^2$A} Instruction Set Architecture. Abbreviations: RF = Register File, CU = Compute Unit, SU = Sample Unit, CDT = Conditional Distribution Table, SE = Sampler Element.
    }
    \label{fig:top_arch}
\end{figure*}
\textbf{3D Visualization:}
As shown in Fig. \ref{fig:3d_roofline}(a), there are three roofs in the \textbf{MC$^2$A} roofline of this CU-SU system, including the SU peak performance, the CU peak performance and the memory peak bandwidth.
Note that these hardware performance boundaries are mutually dependent on each other according to the MCMC algorithm and hardware abstraction (Fig. \ref{fig:3d_roofline}b). 
Therefore, the achievable hardware performance envelope is in the shape of a rectangular frustum. The roof eaves, the CU-performance-bounded corner and the memory-bandwidth-bounded corner cast their polyhedral bottleneck zones.

Given any implementation of a target workload on this hardware, its throughput bottlenecks can be easily traced by using this roofline envelope, either in the 3D or the 2D top-down view.

\textbf{Example Analysis:}
For instance, in an Ising model workload (Fig.~\ref{fig:3d_roofline}c) to update the random variable $0$ with Gibbs sampling, it roughly contains four steps: (1) memory read for its four neighbors' value, (2) 10 operations for energy computation to generate the distribution, (3) then 1 random sampling from the previous generated distribution and (4) update the spin value with the sample result.
We analyze this workload with several hardware configurations in Fig. \ref{fig:3d_roofline}(d). There exists a balanced scenario where the hardware is a perfect match of the algorithm requirements. This golden configuration marks the roofline apex that is highlighted by the purple star in Fig. \ref{fig:3d_roofline}(a). 
For other hardware configurations, the 3D roofline reveals their performance bottlenecks and thus indicates the direction of adjustment.

To sum up, given a target MCMC application and the hardware template topology, our 3D extended roofline model serves as a powerful tool to quickly evaluate the theoretical performance, reveal system-level bottlenecks, and indicate the optimization directions at design time.
These profiling metrics help to boost the MCMC algorithm-hardware co-design flow on choosing optimal accelerator parameters (detailed in Section \ref{sec:result_roofline}).

\section{Accelerator Architecture} \label{sec:architecture}
To leverage the insights from our 3D MCMC roofline profiling and solve the hardware dilemmas in Section \ref{sec:challenges_profiling}, we propose the \textbf{MC$^2$A} hardware architecture with rich flexibility and programmability. 
\textbf{MC$^2$A} is designed to efficiently accelerate a wide range of MCMC algorithms while remaining friendly to scale up or scale out.

We first overview this parameterizable accelerator system in Section \ref{sec:architecture_overview}. 
Next, we elaborate the key components of the \textbf{MC$^2$A} hardware, including the customized instruction set (Section \ref{sec:architecture_isa}), the array of tree-structured Compute Units (Section \ref{sec:architecture_computing_unit}), and the reconfigurable samplers (Section \ref{sec:architecture_sampler}).
Finally, we explain \textbf{MC$^2$A}'s flexible processing data flow and the pipeline schedules of different applications in Section \ref{sec:architecture_processing_flow}.

\subsection{Overview of the Architecture} \label{sec:architecture_overview}
\textbf{MC$^2$A} aims to design a programmable MCMC accelerator, which can support different parallel MCMC algorithms and target general irregular graphs.
Shown in Fig.~\ref{fig:top_arch}a, our accelerator features a hardware loop (HWLOOP) unit in the Instruction Fetch phase to control the step $t$ iteration loop of Alg.~\ref{alg:metropolis-hastings}, a multi-bank register file to support graph irregularity, a pipelined datapath for energy computation and sampling, and the load/store unit for data movement. 
This hardware is scalable by tweaking the design parameters highlighted Fig. \ref{fig:top_arch}(a). Our design choice is powered by the 3D roofline analysis and is detailed in Section \ref{sec:result_roofline}. 

Along with the core modules, \textbf{MC$^2$A} incorporates other modules such as: (1) the multi-bank Register File (RF) to guarantee enough bandwidth in CU-SU data supply and exploit computation-level parallelism; (2) a crossbar-based interconnect to route information between different CU and SU nodes for irregular graph support; (3) the on-chip memory to store input features, weights, intermediate samples and histogram results. These modules are carefully tweaked to have synchronous timing and eliminate internal bottlenecks.

We conduct a comparison with other existing MCMC accelerator architectures.
Compared to chessboard accelerators \cite{spu, pgma}, \textbf{MC$^2$A} is more flexible in supporting the irregular graph with any RV states, such as in advanced Gradient-based discrete sampling algorithms. 
Also, \textbf{MC$^2$A} can accelerate broader MCMC algorithms in parallel than other hybrid systems \cite{aadit2022massively, coopmc, proca, chowdhury2023full}. Moreover, the programmability and highly-efficient hardware design further boost the performance of our accelerator.

\subsection{Instruction Set Architecture (ISA)} \label{sec:architecture_isa}
The programmable hardware is controlled by a custom VLIW (Very Long Instruction Word) instruction set, as shown in Fig.~\ref{fig:top_arch}(c). The ISA includes the following pipeline control types:

\begin{enumerate}
    \item \textbf{Load:} To load data from the on-chip memory into the RF.
    \item \textbf{Compute:} The CU-only mode when the energy computation requires multiple cycles and cannot fit within a single PE (SU bypassed).
    \item \textbf{Sample:} The SU-only mode when sampling the same distribution multiple times (e.g. Step-1 of PAS; CU bypassed). 
    \item \textbf{Compute-Sample:} To execute the energy computation and sampling in a pipelined manner, when the computation can be completed within single instruction.
    \item \textbf{Compute-Sample-Store:} To execute the Compute-Sample and store the results into memory for the final MCMC step.
    \item \textbf{NOP (No Operation):} To resolve pipeline hazards.
\end{enumerate}

During the execution, the instruction is parsed by the pipeline controller to schedule the hardware modules according to the control logic mapping in Fig. \ref{fig:top_arch}(b). 
The bitwidth of each instruction field varies on hardware parameters chosen at the design time, as shown in Fig. \ref{fig:top_arch}(c). We define a dense packing approach for this VLIW ISA to minimize the instruction memory overhead.

\begin{figure}[t]
    \centering
    \includegraphics[width=\columnwidth]{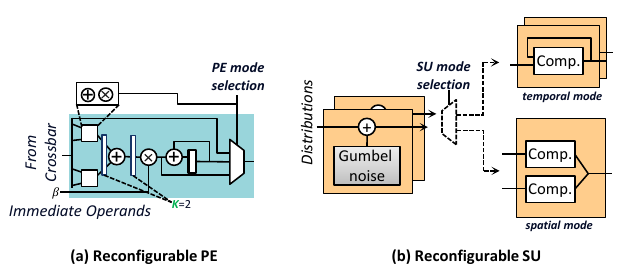}
    \caption{The structure of reconfigurable core elements: (a) Processing element (PE) with different computation functionalities; (b) Samping Unit (SU) with the temporally-iterative and spatially-parallel sampling modes.}
    \label{fig:reconfig_core}
\end{figure}

\begin{figure}[!t]
    \centering
    \includegraphics[trim={0cm 0cm 0cm 0cm}, clip, width=1.0\columnwidth]{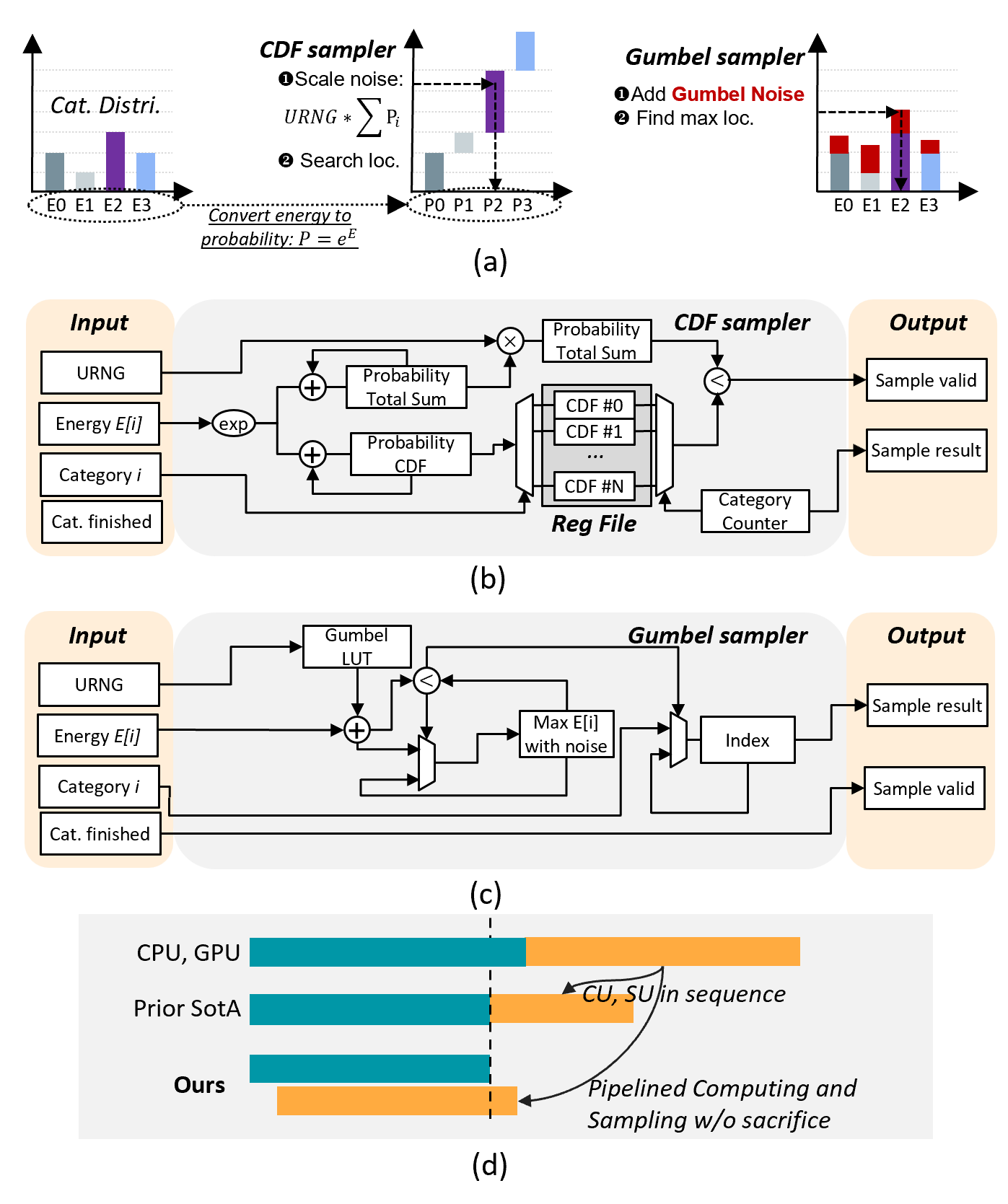}
    \caption{The design comparison of CDF and Gumbel sampler units (SUs): (a) the behavior differences when sampling Categorical Distributions; (b) the architecture of the baseline CDF sampler \cite{spu,pgma}; (c) the architecture of the proposed Gumbel sampler (temporal mode). (d) MC$^2$A's pipelined design improves throughput and latency.
    }
    \label{fig:gumbel_sampler}
\end{figure}   

\subsection{Compute Unit} \label{sec:architecture_computing_unit}
The Compute Unit (CU) consists of $T$ parallel processing elements (PEs) to conduct {\bf Distribution Computing}.

As shown in Fig. \ref{fig:reconfig_core}(a), they form a tree structure to support the reduced-sum or dot-product operation up to the depth of $K$.
The tree is followed by a multiplier and an accumulator to produce inverse temperature $\beta$ and intermediate partial sum. The entire PE array supports a maximum input size of $2^K+1$, including $2^K$ inputs from the RF and one in-place reused intermediate result that reduces the RF access.

The PE can be configured into the following modes:
\begin{enumerate}
    \item \textbf{Bypass:} To bypass the inputs to SU for direct sampling.
    \item \textbf{Dot-Product:} To conduct the energy computation for a random variable $x_i$ in Fig. \ref{fig:algorithm_flow}. 
    \item \textbf{Reduced Sum:} To calculate the reduced-sum in the energy computation for models like the Bayes Nets.
    \item \textbf{Partial Dot-Product or Reduced-Sum:} To partially perform the energy computation for multiple cycles. That is, the \textit{Compute} mode of ISA (SU bypassed).
\end{enumerate}

We cut the PE into $K+1$ pipeline stages to maximize the clock frequency and overall throughput of the system.

\begin{figure*}[!t]
    \centering
    \includegraphics[trim={0cm 0cm 0cm 0cm} , clip, width=1.0\textwidth]{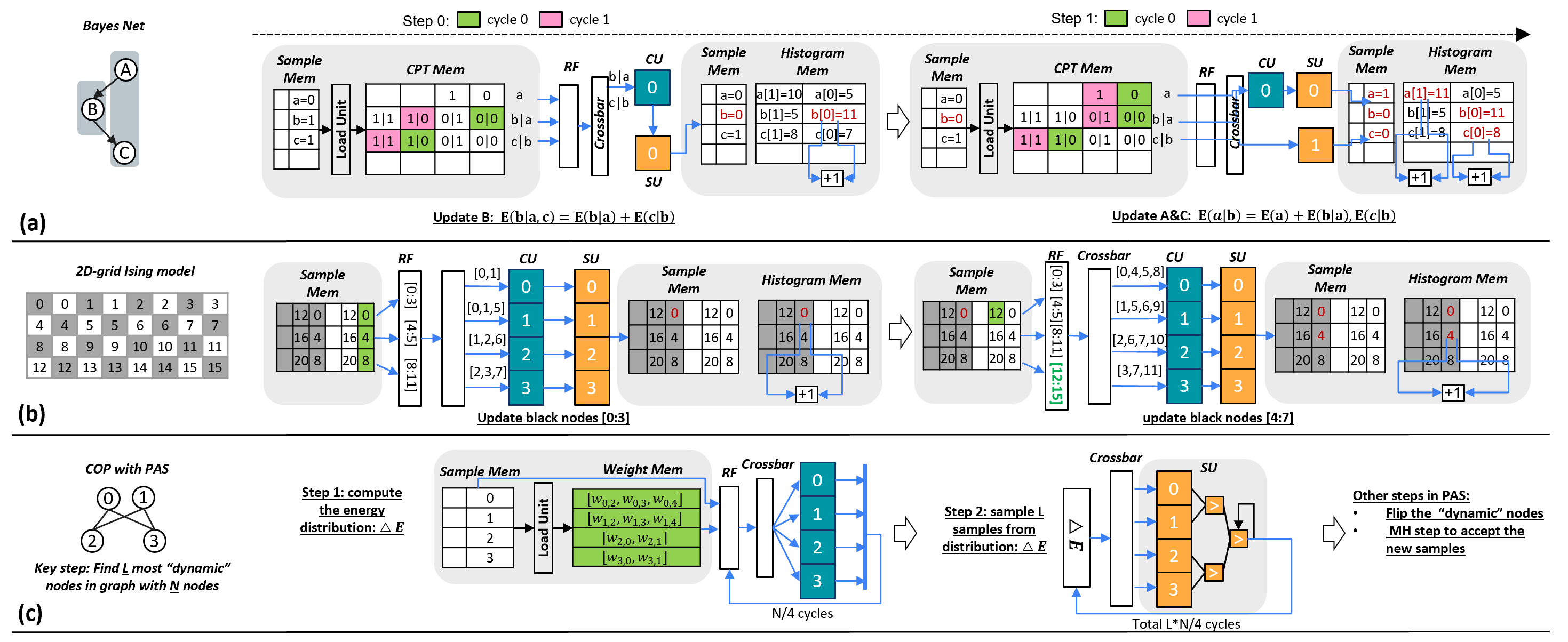}
    \caption{The schedule of flexible processing flows for different MCMC applications: Block Gibbs sampling for (a) Bayes Net and (b) 2D-grid Ising model, and (c) PAS sampling.}
    \label{fig:compute_flow}
\end{figure*}
\subsection{Gumbel-based Sampler Unit} \label{sec:architecture_se}
One core part of our design is the Gumbel-based Sampler Unit (SU) that significantly improves the efficiency. The SU is placed after the CU in MC$^2$A's pipeline. Each SU contains $S$ parallel Sample Elements (SEs) with a depth size of $M$, where $S=2^M$.
The core idea of CDF sampling and Gumbel sampling algorithms is shown in Fig.~\ref{fig:gumbel_sampler} (a).

\textbf{Baseline CDF sampler.} \label{sec:architecture_cdf}
CDF sampler is to sample from a discrete probability distribution using its Cumulative Distribution Function (CDF). Given probabilities $[p_0, ..., p_n]$ where $\sum P_i = 1$. The CDF sampler contains following steps:
\begin{enumerate}
    \item Compute the CDF, where $CDF = [p_0, p_0+p_1,..., \sum p_i]$.
    \item Generate the uniform random number $u \sim Uniform(0,1)$.
    \item Find the smallest index $i$ such that $u < CDF[i]$.
    \item Return $i$ as the sample result.
\end{enumerate}
For un-normalized energy inputs, the CDF sampling requires exponential and normalization operations to generate the probability distribution. To simplify the operations, in practice, the hardware CDF sampler (shown in Fig.~\ref{fig:gumbel_sampler}(b)) using the “URNG * Total Sum” to scale the range of uniform random number from [0-1] to [0 - $\sum p_i$], then compare it with each probabilities.

\textbf{Gumbel-max trick and hardware implementation.} \label{sec:architecture_sampler}
The Gumbel-max trick provides an efficient method to sample from a discrete distribution. It leverages the property that if $X_i$ is a random variable with a discrete distribution, then $-\log(-\log(X_i))$ follows a Gumbel distribution. The trick works as follows for a Categorical distribution with probabilities $\{p_1, p_2, ..., p_n\}$ and $j\in [1,n]$:
\begin{enumerate}
    \item Generate $u_j$ for each category $j$ from a uniform distribution.
    \item Compute $-\log(-\log(u_j))$ to obtain Gumbel noise.
    \item Add the Gumbel noise to the log-probabilities.
    \begin{equation}
        g_j = -\log(-\log(u_j)) + \log(p_j)
    \end{equation}
    \item Select the category with the maximum value: $j = \arg\max_j g_j$.
\end{enumerate}

Our Gumbel sampler is built with an LUT for converting uniform noise to Gumbel noise and a comparator for sequentially selecting the maximum value (Fig.~\ref{fig:gumbel_sampler}c). 

\textbf{Benefits.}
As shown in Fig.~\ref{fig:gumbel_sampler}, the Gumbel sampler offers several advantages over the baseline CDF sampler:
\begin{enumerate}
    \item It directly processes unnormalized energies, eliminating \textit{exp} and \textit{norm} operations required by the CDF sampler.
    \item It reduces time complexity by \textbf{$2\times$} (shown in Fig.~\ref{fig:gumbel_sampler}(d)). The CDF sampler requires $O(2\mathcal{N}+1)$ operations to compute the Cumulative Distribution Table (CDT) and search for the sampling index, while the Gumbel sampler achieves $O(\mathcal{N})$ complexity with the pipeline design.
    \item It simplifies the hardware without the need of an internal register file for the CDT.
    \item It leverages logarithmic operations to compute the energy function thus avoiding underflow and overflow issues \cite{sountsov2024running}.
\end{enumerate}

\textbf{Reconfigurability.}
As shown in Fig. \ref{fig:reconfig_core}(b), the SU can be configured to operate in either \textbf{\textit{temporal}} or \textbf{\textit{spatial}} mode, depending on workload requirements. In \textit{temporal mode}, a single comparator per input is used to find the maximum value index iteratively over ${N}$ cycles for a size-$N$ distribution. In \textit{spatial mode}, the SU combines multiple comparators into a 2D tree structure to sample multiple distributions concurrently in a single cycle.

\subsection{Processing Flow} \label{sec:architecture_processing_flow}
With the programmable CU and SU organized in a pipelined data path, \textbf{MC$^2$A} is flexible enough to process different MCMC applications with ultra-high utilization.

\textbf{Basic Data Flow Pipelining.}
Unlike other MCMC accelerators \cite{coopmc, proca, pgma, spu, shiruizhaoAIA16nmMulticore2024} that accelerate the sampling procedure with a discrete hardware sampler, the CU and SU in \textbf{MC$^2$A} (Fig. \ref{fig:top_arch}(a) has internal pipelines. Once the CU finishes the energy computation of one distribution bin, the SU starts directly to process the energy with Gumbel noise insertion. 
Therefore, the idle period from the computing-sampling dependency is hidden by the pipelining so that the hardware utilization is maximized. 
The only exception is the \textit{Compute} and \textit{Sample} type in pipeline control (Section \ref{sec:architecture_isa}), which means that the workload exceeds the capacity of CU and SU. Then \textbf{MC$^2$A} falls back into a multi-cycle processing schedule.
Note that the CU-SU capacity is a design-time decision (Section \ref{sec:result_roofline}), thus complementary but orthogonal to the run-time pipelining here.

\textbf{Application-specific Workflow.}
To demonstrate the flexibility and strength of \textbf{MC$^2$A}, we present the processing flows for three application examples: Block Gibbs sampling applied to (A) a Bayes Net and (B) an Ising model, as well as (C) the PAS sampling for COP.
Their processing schedules are illustrated in Fig.~\ref{fig:compute_flow}(a), (b), and (c), respectively.
The hardware configuration used in these examples is $S=T=4, K=1, B=12$. For workloads with massive RV-level parallelism, we use the temporal-mode SU. For single distribution cases, we switch to the spatial-mode SU to improve the latency.

\textbf{(A) Bayes Net with Temporal-Mode SU.}
In the given toy example of Fig.~\ref{fig:compute_flow}(a), each RV has 2 possible sample values, and their conditional probabilities are stored in different memory banks. The CPT of RV A is stored in the first row in CPT memory. We use $0$ to represent $P(a=0)$. For RV B, $1|0$ is $P(b=1|a=0)$. Due to RV-parallelism (Fig. \ref{fig:parallelism}), RV-$B$ can update at step-0 where $P(B|A,C)\sim P(B|A)P(C|B)$. We compared the energy with the addition and sampling in pipeline way. RV-$A$ and $C$ update in parallel at step-1 in a similar way. The intermediate results are stored in different RF banks to avoid conflicts. For instance, it takes two instruction cycles to update RV-$B$:
\begin{enumerate}
    \item \textbf{cycle 0:} \textbf{Load} the CDT memory of $E(b=0|a=0)$ and $E(c=1|b=0)$, according to the current sample memory. The crossbar connects the RF values to CU-0 and SU-0 for the computation of $E(b=0|a=0,c=1)$.
    \item \textbf{cycle 1:} \textbf{Load} the CDT memory of $E(b=1|a=0)$ and $E(c=1|b=1)$, according to the current sample memory. Then $E(b=1|a=0,c=1)$ is computed and compared to the cycle-0 $E(b=0|a=0,c=1)$. Then the sample and histogram memory content is updated based on the comparison result.
\end{enumerate}

\textbf{(B) 2D-grid Ising Model with Temporal-Mode SU.}
For this structured graph (Fig.~\ref{fig:compute_flow}(b), same setup as \cite{yang2019high}), we can divide the workload into two categories in a chessboard way. The number shows in sample memory the RV id. To update $black RV3$, it need to access its three $white$ neighbors ($white$ $RV2,3,7$). The formula shown in Fig.~\ref{fig:algorithm_flow} is used to compute the total energy for the current $black RV3$. All $black$ and $white$ nodes can be updated in the same step theoretically. Constrained by the hardware resource, we update 4 RVs to fully utilize the CU and SU. For instance, in step 0, $black$[0:3] are updated by accessing $white$[0:11].  
The crossbar guarantees the maximal RF data reuse across CUs.
Compared to Bayes Nets, there is no indirect CDT memory access pattern, thus a higher hardware utilization because of more parallelizable RVs.

\textbf{(C) PAS for COPs and EBM with Spatial-Mode SU.}
The key step in the PAS sampling algorithm is to find the top $L$ most "dynamic" RVs, which requires sampling from a large distribution (size of 125 $\sim$ 1347) for at least $L$ times, which is challenging in terms of high complexity.
In \textbf{MC$^2$A}, we avoid the CDT register file overflow (e.g. in \cite{spu, pgma}) with multi-cycle \textit{\textbf{Compute}} and \textit{\textbf{Sample}} pipeline controls from Section \ref{sec:architecture_isa}. We further configure the SU into \textit{spatial mode} to improve the latency performance. 
As shown in Fig.~\ref{fig:compute_flow}(c), it only requires $N/4$ cycles to finish the distribution $\Delta E$ computing and then $L*N/4$ cycles to sample the RV indexs $J$, where $N$ is the distribution size.

To sum up, the efficient processing flow of all three applications above showcases the efficacy and flexibility of our \textbf{MC$^2$A} architecture. More benchmark evaluations are in Section \ref{sec:result}.

\section{Experiments and Discussions} \label{sec:result}
To verify the strength of the \textbf{MC$^2$A} co-design workflow for MCMC applications, we evaluate our system with a set of benchmarks. In this section, we first introduce the experimental setups and benchmark applications in Section \ref{sec:result_setup}. 
Then, in Section \ref{sec:result_roofline}, we showcase the usage of our 3D MCMC Roofline model to quickly determine optimal design parameters for the \textbf{MC$^2$A} accelerator. 
Next, we conduct an ablation study in Section \ref{sec:result_ablation_SU} for the performance and accuracy of our Gumbel Sampler Unit. 
Finally, we evaluate the system performance of \textbf{MC$^2$A} and compare it with the state-of-the-art in Section \ref{sec:result_sota_comparison}.

\subsection{Experimental Setup} \label{sec:result_setup}
We compose our benchmark set from representative MCMC algorithms and applications, including Bayes Nets from the Bayes Net repository \cite{bn_repository}, MRF workloads from \cite{spu}, as well as Combinatorial Optimization Problems (COP) and energy-based models from the DISCS benchmark suite \cite{katayoongoshvadiDISCSBenchmarkDiscrete2023}. A detailed summary of the workloads and their associated MCMC algorithms is in Tab.~\ref{tab:workloads}.

\begin{table*}[!t]
    \centering
    \small
    \caption{\textbf{Workloads for experiments} adapted both from \cite{box2011bayesian} and \cite{katayoongoshvadiDISCSBenchmarkDiscrete2023}, including Bayes nets, MRF/Ising model, Combinatorial Optimization Problems (COP) and energy-based model.}
    \begin{NiceTabular}{cccccc}
    \hline
     Name & Model & Application & Nodes & Edges & Algorithm \\
    \hline
        Earthquake & Bayes Net & models the probability of an earthquake occurring & 5 & 4 & BG \\ 
        Survey & Bayes Net & models student grades, intelligence, and difficulty relationships & 6 & 6 & BG \\ 
        Image Seg. & MRF/Ising & using MRF to perform image segmentation & 150k & 600k & BG \\ 
    \midrule
        ER700 & MIS & Maximum Independent Set. The graphs are from Satlib & 1347 & 5978 & PAS \\ 
        Twitter & Max clique & Maximum subset of vertices, all adjacent to each other. & 247 & 12174 & PAS \\ 
        Optsicom & MaxCut &  Partition vertices into two sets to maximize edge cuts. & 125 & 375 & PAS \\ 
    \midrule
        RBM & EBM & Binary RBM with hidden dimension 25  &  809 & 19K & PAS \\ 

    \bottomrule
    \end{NiceTabular}
    \label{tab:workloads}
\end{table*}     

\textbf{Baseline Designs.}
we consider the RTX GPU, Xeon CPU, and custom ASIC accelerators (SPU\cite{spu}, PGMA\cite{pgma}, CoopMC\cite{coopmc}, sIM\cite{aadit2022massively}) as our baseline platforms.

\textbf{Software Setups.} For Bayes Nets, we utilize the pyAgrum framework \cite{ducamp2020agrum} to benchmark Gibbs sampling on the Xeon CPU and Baylib \cite{bayeslib} on the GPU. For the remaining workloads, we employ JAX-based implementations from \cite{katayoongoshvadiDISCSBenchmarkDiscrete2023} on both the CPU and GPU.

\textbf{Hardware Setups.}
To evaluate the energy and area of our design, we implement the accelerator in RTL using SystemVerilog and synthesize it using Cadence Genus. The design targets an Intel $16nm$ process node with a clock frequency of $500$ MHz. A cycle-accurate simulator is developed to profile the accelerator.

\begin{figure}[t]
    \centering
    \includegraphics[width=\columnwidth]{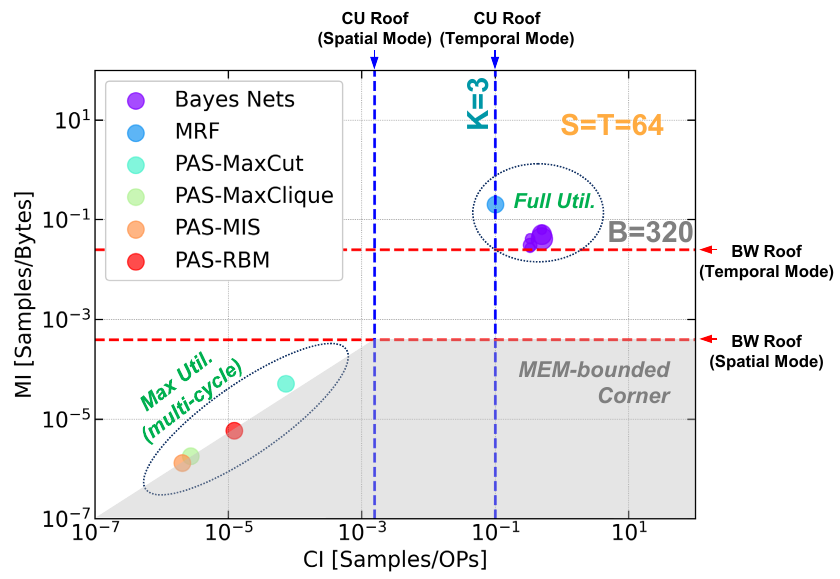}
    \caption{The determination of hardware design parameters from algorithm-hardware co-design based on 3D roofline model analysis (top-down view). The dual roofline groups result from \textbf{MC$^2$A}'s spatial/temporal-mode reconfigurability.
    }
    \label{fig:workload_roofline}
\end{figure}

\subsection{Rapid Co-design with 3D MCMC Roofline} \label{sec:result_roofline}
We determine the hardware parameters of the \textbf{MC$^2$A} accelerator by evaluating key workloads in our benchmark set (Tab.~\ref{tab:workloads}), including one Bayes Nets, MRF, three COP problems, and one EBM (RBM). The procedure of design parameter choice is shown in the top-down view of the 3D roofline model in Fig.~\ref{fig:workload_roofline}. 

Firstly, we attempt to avoid either the data memory bottleneck (gray region in Fig.~\ref{fig:workload_roofline}) and the hardware resource outage. This is extremely critical for PAS workloads. Based on Section \ref{sec:architecture_processing_flow} and the total hardware resource budget (the scale of SU throughput, i.e. the z-axis height in the 3D roofline), we choose the single/multiple-cycle CU modes and the spatial/temporal SU modes for each workload and plot their locations onto the 3D roofline. From our budget, we choose the size of CU and SU to be $S=T=64$.

Secondly, after the memory bottleneck is cleared. We scale the \textbf{MC$^2$A} accelerator to fit these scheduled workloads. For those temporal-mode workloads with massive parallelism (Bayes Nets and MRF), we pursue the full utilization of the CU-SU system. For those multi-cycle workloads (PAS series, CU-bound), we try to minimize the idle cycles of SU by pushing the spatial-mode roof apex point towards these workloads as much as possible. 
Therefore, we set the CU depth with $K=3$ and the memory bandwidth to $B=320$.

Finally, we calculate all the correlated hardware design parameters. In all, we get the following hardware parameters for evaluation: CU size of $T=64$ and $K=3$, SU size of $S=64$ and $M=6$, and memory bandwidth $B=320$. The data width of each input data memory bank is 32 bits to support floating-point precision weights and CPTs (stored in their logarithmic values for logarithmic computation). The basic memory IP block, generated by the Intel memory compiler, is 8KB with a size of 1024$\times$32 bits. To support a maximum distribution size of 256 and a chain length of $10^6$, the total number of memory IP blocks is 600 ($320$ for input data memory, $320*8/32$ for sample memory, and $320*20/32$ for histogram memory), resulting in a total on-chip SRAM size of $4.8$ MB.

Therefore, by leveraging the proposed 3D MCMC roofline model, we enable a straightforward and general approach to solve the complicated MCMC accelerator design challenges with promising hardware profiles (further evaluations in Section \ref{sec:result_sota_comparison}).

\begin{figure}[!t]
    \centering
     \includegraphics[width=0.9\columnwidth]{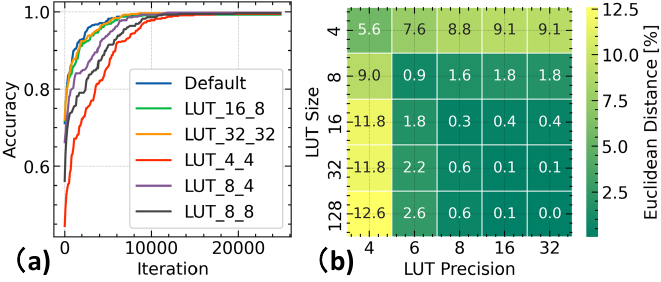}
    \caption{The accuracy performance of different Gumbel LUT size and precision over (a) real workload: MaxCut problem and (b) randomly generated distribution, where size-16 LUT and 8-bit precision provides good-enough accuracy.
    }
    \label{fig:su_accuracy}
\end{figure}

\begin{figure}[!t]
    \centering
    \includegraphics[width=0.9\columnwidth]{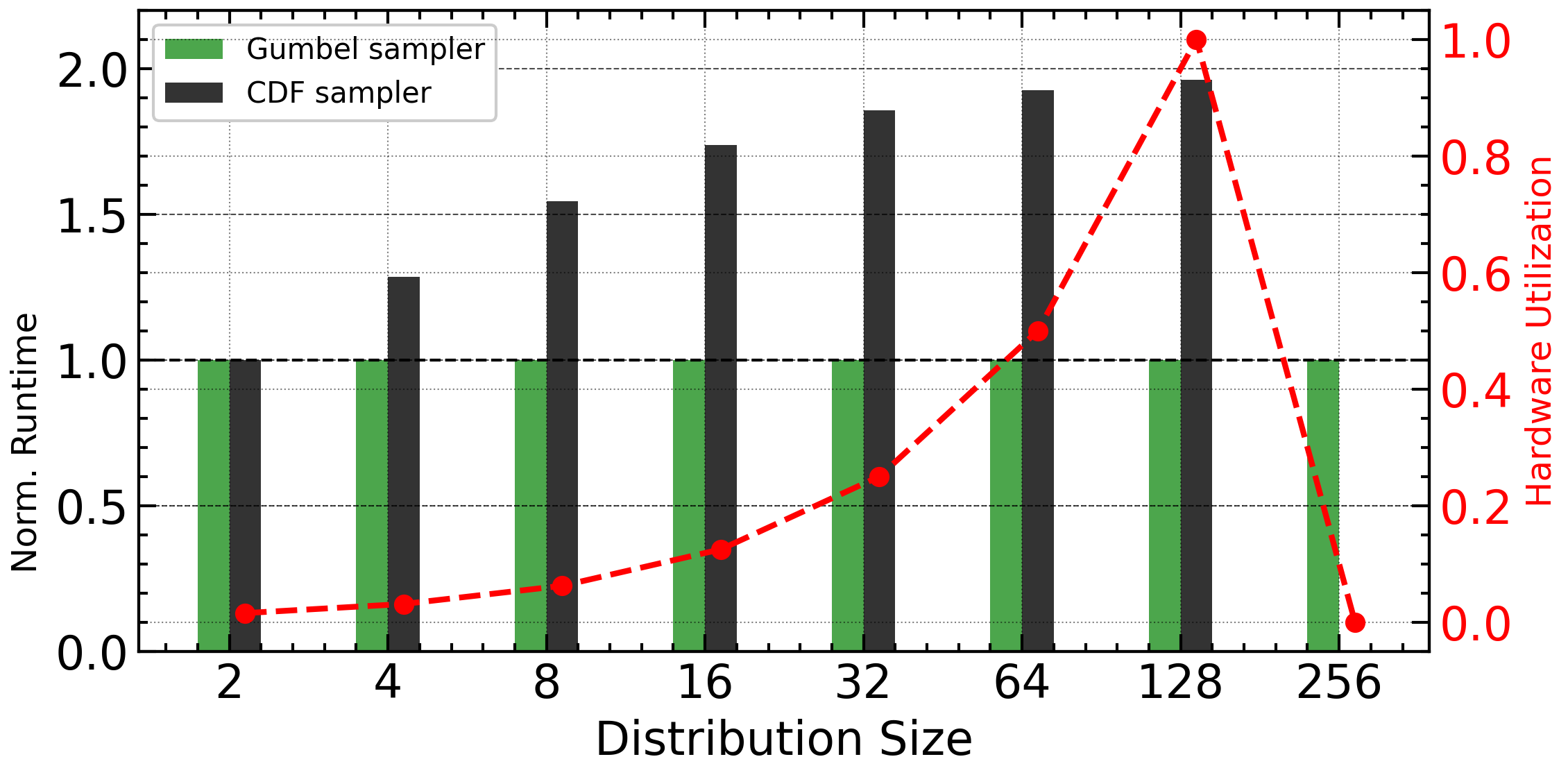}
    \caption{Comparison between Gumbel sampler and traditional CDF sampler. CDF sampler always requires more runtime due to the sequential nature, and the hardware utilization drops when increasing the distribution size and fails at size-256.  
    }
    \label{fig:gumbel_comparison}
\end{figure}

\subsection{Optimal SU Design} \label{sec:result_ablation_SU}
To evaluate the Gumble lookup table (LUT) size required for the generation of Gumbel noise, 
we conduct an ablation study on the accuracy performance of the Gumbel LUT with different sizes and precisions. The results are shown in Fig.~\ref{fig:su_accuracy}(a) for the MaxCut problem and (b) for 100 random distributions sampled $10^6$ times. The analysis indicates that a LUT size of 16 and a precision of 8 provide sufficient accuracy performance for various workloads.

In Fig.~\ref{fig:gumbel_comparison}, we compare the throughput of the Gumbel sampler with the traditional CDF sampler used in SPU \cite{spu}, PGMA\cite{pgma}. The results demonstrate that the hardware utilization of the CDF sampler decreases with increasing distribution size, limiting its support for larger distributions. Further, large distributions with a range size bigger than the design scope are not supported by CDF-based hardware \cite{spu,pgma,coopmc}.
In contrast, the Gumbel sampler maintains consistent throughput across different distribution sizes, showcasing its flexibility and efficiency. 

\begin{figure}[!t]
    \centering
    \includegraphics[width=\columnwidth]{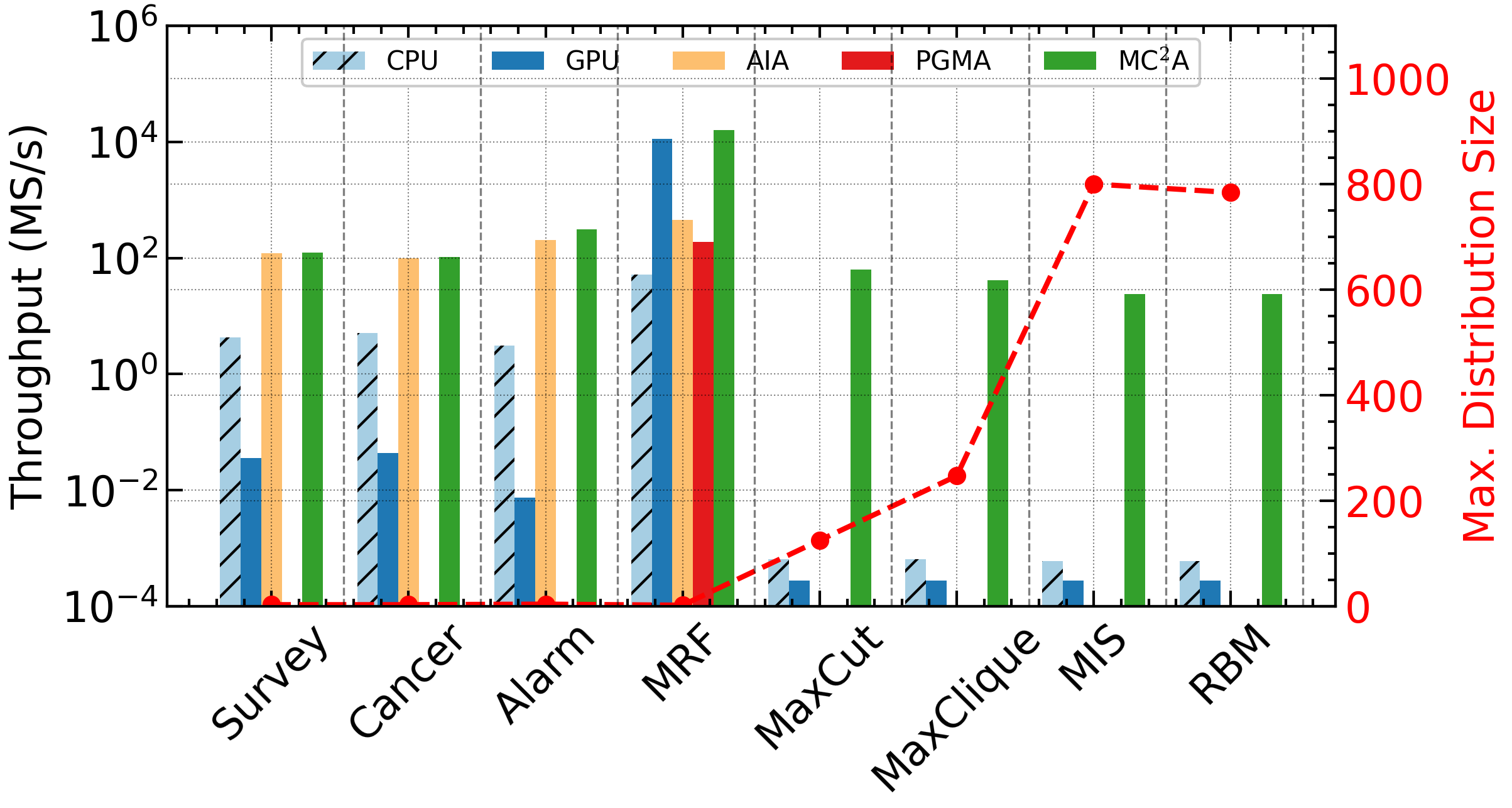}
    \caption{Latency comparison. MC$^2$A works under \textit{temporal mode} for Bayes Net and MRF. For the key step in PAS sampling algorithm where the SU is under \textit{spatial mode}. Compared to other computing platforms, MC$^2$A is shown higher throughput and flexibility for various workloads.}
    \label{fig:throughput_performance}
\end{figure}
\begin{figure}[!t]
    \centering
    \includegraphics[width=1\columnwidth]{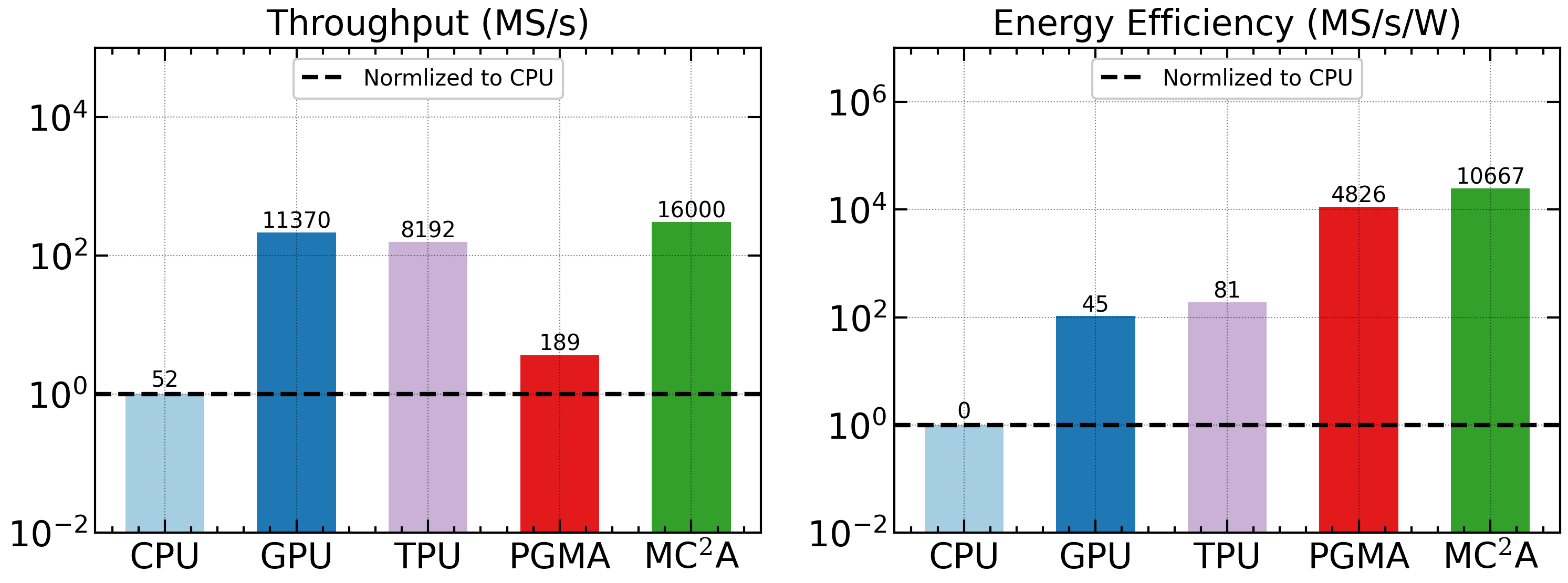}
    \caption{Energy Efficiency Improvement on structured graph. MC$^2$A consistently improves energy efficiency, compared to CPU, GPU, TPU and existing accelerator.}
    \label{fig:energy_performance}
\end{figure}

\subsection{State-of-The-Art Comparison} \label{sec:result_sota_comparison}
We compare the throughput performance results of MC$^2$A in terms of throughput (GS/s) and latency for various workloads, including Bayes Nets, MRF, COP and RBM, against the state-of-the-art solutions of Xeon CPU, RTX 2080Ti GPU and ASIC accelerators in Fig.~\ref{fig:throughput_performance}. 
For fair comparison, we used same data type (int 32bit) as same as PGMA\cite{pgma} in the CU. The other hardware configuration paramters are following the description in Sec.\ref{sec:result_roofline}. All results are based on single chain acceleration.
Our \textbf{MC$^2$A} consistently outperforms other existing hardware platforms across all workloads. 

\subsubsection{Compared with CPU, GPU and TPU}
\textbf{Irregular Bayes Nets.}
The throughput performance of MC$^2$A is significantly higher than that of the CPU and GPU platforms for Bayes Nets (Survey, Cancer, Alarm). The average throughput improvement is 25$\times$ and over 1e3$\times$ compared to CPU and GPU, respectively.
Powerful GPUs exhibit worse throughput. This is due to the following reasons:
\ding{172}\textbf{Inefficient handling of irregular data access:} The indirect memory access patterns in Bayes Nets and the irregular graph structures in COPs result in poor GPU utilization.
\ding{173}\textbf{High kernel launch and data movement overhead:} MCMC workloads involve Pseudo-Random Number Genration (PRNG) and random sampling from distribution; these small, quick tasks can not be executed fast on GPU, which prefers bulk operations.

\textbf{Structured Graphs.}
For structured graphs like 2D-grid MRF, the GPU and TPU shows better performance than the CPU. This is because the GPU and TPU can exploit the massive parallelism in MRF workloads, where multiple RVs can be updated simultaneously. MC$^2$A provides even better performance due to dedicated sampler design and pipelined CU/SU schedule.  MC$^2$A achieves 307.7$\times$, 1.4$\times$ and 2.0$\times$ better performance than CPU Xeon, Nvidia Tesla V100 GPU, Google TPU-v3 core.

\textbf{COP and EBM}
For three COP workloads and one EBM, we learned the performance of key operations in PAS sampling (shown in Fig.~\ref{fig:compute_flow}(c)). The throughput performance of MC$^2$A is significantly higher than that of CPU and GPU platforms. The average latency improvement is 60$\times$ and 142$\times$ compared to CPU and GPU, respectively. We also observed our latency increases with the distribution size, which is due to the increased number of cycles required for both sampling and computing.

\textbf{Energy Efficiency Comparison.}
The energy efficiency of MC$^2$A is evaluated in terms of Giga-Samples per second per Watt. The results are shown in Fig.~\ref{fig:energy_performance}. 
Overall, MC$^2$A demonstrates significant energy efficiency improvements compared to CPU, GPU and TPU platforms (with TDP of 120W, 250, 100W), with an average reduction of 10K$\times$, $355\times$, and $197.5\times$, respectively. 

\subsubsection{Compared to SotA accelerators}
SPU\cite{spu}, PGMA\cite{pgma} were dedicated design just for MRF with a fixed datapath. CoopMC\cite{coopmc} was a similar hardware framework as MC$^2$ with the optimized tree CDF sampler. sIM\cite{aadit2022massively} is a customized design only for Ising models (RV states are equal to 2). PROCA\cite{proca} utilized the In-Memory Computing unit for sampling. However, they show lower throughput performance due to the inefficient \textit{sequential} CU and SU design.
Compared to the reported throughput performance, MC$^2$A achieves 4.8$\times$, 84.2$\times$, 32$\times$ and 80$\times$ better performance than SPU\cite{spu}, PGMA\cite{pgma},  CoopMC\cite{coopmc} and PROCA\cite{proca} respectively.

PROCA \cite{proca} and our design are the only two designs that support \textit{any} distribution size. However, PROCA spends more area to update one RV, due to each PROCA core corresponds to one RV, and it requires the vector RISC-V for computation, which shows less area efficiency.

\section{Conclusion}\label{sec:conclusion}

This paper introduces \textbf{MC$^2$A}, a novel algorithm-hardware co-design framework for accelerating MCMC applications. Key contributions include a 3D MCMC roofline model for workload profiling and rapid HW-SW co-design, a flexible and programmable hardware architecture, and a Gumbel-based sampler for efficient sampling.

The resulting hardware accelerator outperforms CPUs, GPUs, and other existing MCMC accelerators in performance and energy efficiency. Evaluated on diverse MCMC algorithms and applications, \textbf{MC$^2$A} is proved to enable broader adoption of MCMC methods in real-world scenarios.


\bibliographystyle{IEEEtran}
\bibliography{refs}

\clearpage

\end{document}